\documentclass[10pt,journal,compsoc]{IEEEtran}

\ifCLASSOPTIONcompsoc
  \usepackage[nocompress]{cite}
\else
  \usepackage{cite}
\fi

\usepackage{xcolor}
\usepackage[normalem]{ulem} 
\newcommand\hl{\bgroup\markoverwith
{\textcolor{yellow}{\rule[-.5ex]{2pt}{2.5ex}}}\ULon}

\usepackage{amsmath,amsfonts}
\usepackage{algorithmic}
\usepackage{algorithm}
\usepackage{array}
\usepackage{subfig}
\usepackage{textcomp}
\usepackage{stfloats}
\usepackage{url}
\usepackage{verbatim}
\usepackage{graphicx}
\usepackage{cite}
\usepackage{diagbox}
\usepackage{algorithm}
\usepackage{algorithmic}
\usepackage{wasysym}
\usepackage{hyperref}

\ifCLASSINFOpdf
\else
\fi

\begin{document}
%
\title{Revisiting Few-Shot Learning from a Causal Perspective}
%
%
%
%

\author{    Guoliang Lin,
    Yongheng Xu,
   Hanjiang Lai,
   Jian Yin
\IEEEcompsocitemizethanks{\IEEEcompsocthanksitem G. Lin and Y. Xu are with School of Computer Science and Engineering, Sun Yat-Sen University, China. \protect \\ 
E-mail: \{lingliang, xuyh59\}@mail2.sysu.edu.cn. 
\IEEEcompsocthanksitem H. Lai is with School of Computer Science and Engineering, Sun Yat-Sen University, China, and also with Guangdong Key Laboratory of Big
Data Analysis and Processing, Guangzhou, China. Hanjiang Lai is Corresponding author. \protect \\
E-mail: laihanj3@mail.sysu.edu.cn. 

\IEEEcompsocthanksitem J. Yin is with School of Artificial Intelligence, Sun Yat-Sen University, China, and also with Guangdong Key Laboratory of Big
Data Analysis and Processing, Guangzhou, China.\protect \\
E-mail:issjyin@mail.sysu.edu.cn.}
\thanks{This work is supported by the National Natural Science Foundation of China (U22B2060, U1911203, U2001211), Guangdong Basic and Applied Basic Research Foundation (2019B1515130001, 2021A1515012172), Key-Area Research and Development Program of Guangdong Province ( 2020B0101100001).}
}

%
%

\markboth{IEEE Transactions on Knowledge and Data Engineering,~Vol.~14, No.~8, August~2023}%
{Lin \MakeLowercase{\textit{et al.}}: Revisiting Few-Shot Learning from a Causal Perspective}
%



\IEEEtitleabstractindextext{%
\begin{abstract}
 Few-shot learning with $N$-way $K$-shot scheme is an open challenge in machine learning. Many metric-based approaches have been proposed to tackle this problem, e.g., the Matching Networks and CLIP-Adapter. Despite that these approaches have shown significant progress, the mechanism of why these methods succeed has not been well explored. In this paper, we try to interpret these metric-based few-shot learning methods via causal mechanism. We show that the existing approaches can be viewed as specific forms of front-door adjustment, which can alleviate the effect of spurious correlations and thus learn the causality. This causal interpretation could provide us a new perspective to better understand these existing metric-based methods.  Further, based on this causal interpretation, we simply introduce two causal methods for metric-based few-shot learning, which considers not only the relationship between examples but also the diversity of representations. Experimental results demonstrate the superiority of our proposed methods in few-shot classification on various benchmark datasets. Code is available in \href{https://github.com/lingl1024/causalFewShot}{https://github.com/lingl1024/causalFewShot}. 
\end{abstract}

}

\maketitle

\IEEEdisplaynontitleabstractindextext

%
\IEEEpeerreviewmaketitle

\ifCLASSOPTIONcompsoc
\IEEEraisesectionheading{\section{Introduction}\label{sec:introduction}}
\else
\section{Introduction}
\label{sec:introduction}
\fi

%
%
%
%
\IEEEPARstart{D}{ata} plays an important role in machine learning. To learn good machine learning models, such as classification~\cite{he2016deep, dosovitskiy2020image}, supervised training on these tasks always requires large amounts of data, especially for deep learning. But it is hard to obtain sufficient labeled training data in most cases. When the available training data is very rare, inspired by that humans can make remarkably good predictions just from few observations, \textbf{few-shot learning} (FSL)~\cite{wang2020generalizing,han2023towards} has emerged as an appealing approach. 
In this paper, we focus on the $N$-way $K$-shot classification task~\cite{vinyals2016matching,snell2017prototypical}, where there are $N$ classes and each class has only $K$ labeled training examples.

In recent years, considerable research effort has been devoted to developing efficient algorithms for FSL, e.g., the meta-learning methods~\cite{finn2017model, yao2019automated, wang2022global}. 
Among these methods, a notable research line is the metric-based methods~\cite{li2019finding,zhou2022forward,liu2022dynamic}, which perform the distance computations between the test example and training examples in a learned metric space. Then, the core idea behind metric-based few-shot methods is to assign the label according to the distance computations. For example, Matching Networks~\cite{vinyals2016matching} proposed a weighted nearest neighbour classifier, where a neural network is employed to learn an embedding function and then the distances are computed based on these deep neural features. After that, the label of test example can be  obtained via the distance computations between this test example and all $N \times K$ training examples. Similar to that, Prototypical Networks~\cite{snell2017prototypical} performed classification by computing distances between the test example and the class prototypes of training examples, where the class prototype is the mean of the representations of the training examples belonging to this class. 

Further, with the rapid development of large-scale pre-trained models, metric-based methods~\cite{Zhang_2023_CVPR,zhang2022tip} also have been proposed to exploit the powerful feature representations extracted from these pre-trained models. For example, the large-scale pre-trained vision-language models, e.g., CLIP~\cite{radford2021learning}, have also shown surprising results for FSL. The core component is that these vision-language pre-trained models directly learn to align images with raw texts, thus additional image and text distance information is available. For example, CLIP-Adapter~\cite{gao2021clip} was based on the pre-trained model CLIP~\cite{radford2021learning}, and it fine-tuned feature adapters with the help of the language branch. Tip-Adapter-F~\cite{zhang2022tip} further improved CLIP-Adapter by acquiring the weight of the feature adapter from a cache model constructed by training data. \textbf{Despite the success of these methods, our theoretical understanding of this success remains limited.} 

\begin{figure}[t]
    \centering
    \includegraphics[scale=0.15]{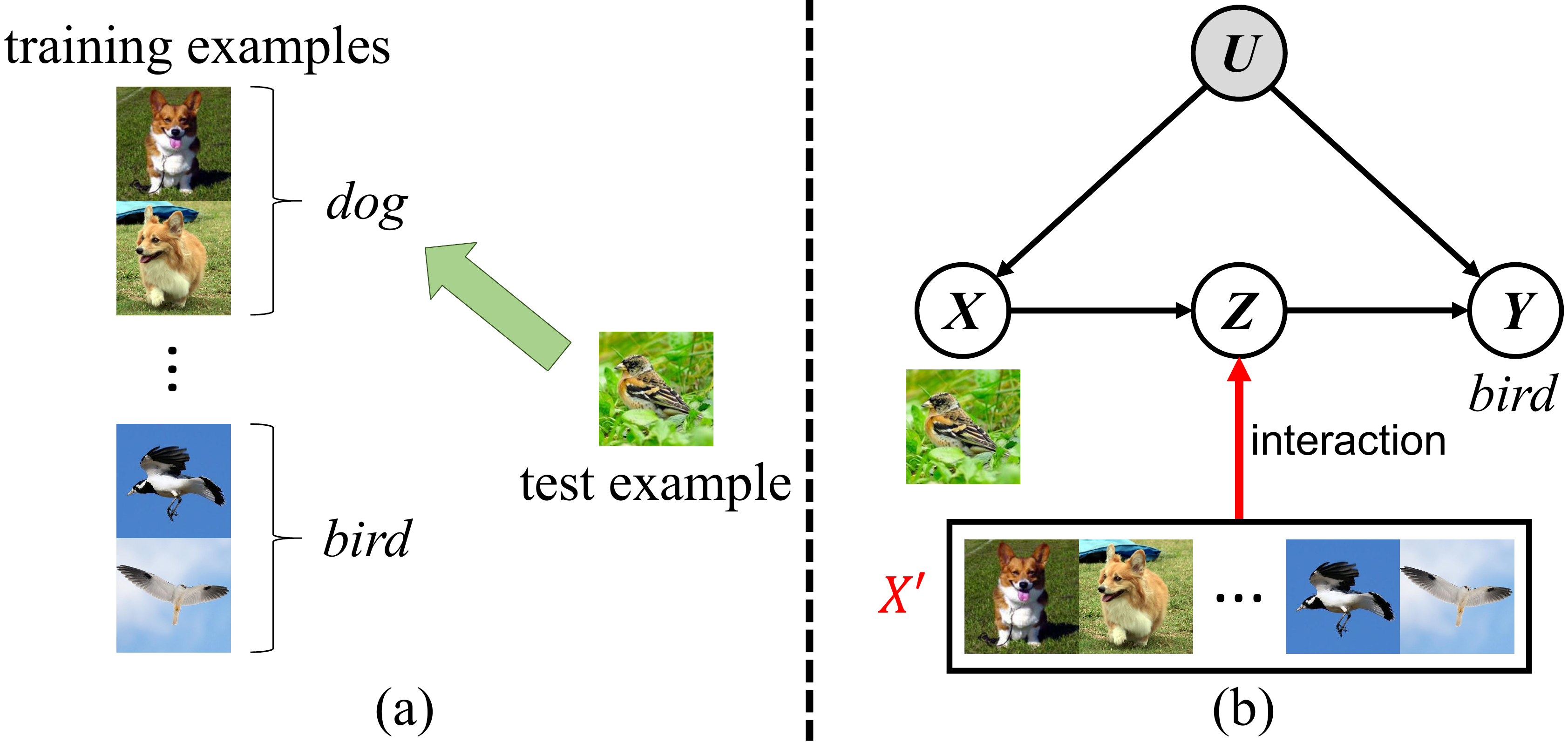}
    \caption{(a) The unobserved confounders, e.g., ``taking pictures of dogs in the grass'', would mislead the training model to learn spurious correlation: the model would tend to classify grass as the dog. This is incorrect for test examples of birds when it is also ``taking pictures of birds in the grass". (b) The causal graph for better understanding the existing few-shot learning methods from the perspective of removing confounding factors by the front-door adjustment. $U$: unobserved confounder, $X$: example, $Z$: representation of the example, $Y$: label. }
    \label{causal}
\end{figure}

Recently, causal inference~\cite{zhang2020causal, tang2020long} has attracted much attention, which aims to seek causality behind any association~\cite{glymour2016causal}. One very common situation~\cite{li2021confounder,huang2022deconfounded,yang2021deconfounded} that causes spurious correlations between examples $X$ and labels $Y$ is shown in Figure~\ref{causal} (b), where $X \to Y$ is also affected by the \textbf{unobserved confounders} $U$~\cite{glymour2016causal,huang2022deconfounded}.For example, $X$ is a picture of a bird in the sky, $Y$ is the semantic class of birds and $U$ denotes ``taking pictures of birds in the sky''. The $U$ is the way how we collect and construct the training dataset. When we train a deep model, we do not know $U$ and only the training samples $X$ and $Y$ are available. Since deep models always learn the co-existing correlations between the training samples and labels, the $X \leftarrow U \to Y$ would mislead the training model to learn the spurious correlation between $X$ and $Y$, e.g., $sky \to bird$. 

In few-shot learning tasks, this problem is especially serious since there are only a few training examples which can lead to data bias more easily. As illustrated in Figure~\ref{causal} (a) for a few-shot task, the unobserved confounder, such as ``taking pictures of birds in the sky'', would mislead the training model to learn spurious correlation $sky \to bird$. It may lead to the misclassification of examples from a new distribution, e.g., ``taking pictures of birds in the grass". Such spurious correlation is harmful for few-shot learning~\cite{luo2021rectifying}.  To block the paths from the confounders, \textit{back-door adjustment}~\cite{glymour2016causal} and \textit{front-door adjustment}~\cite{glymour2016causal} are two popular approaches. The front-door adjustment does not require any knowledge of the confounders, thus it is an appealing approach to deconfound the unobserved confounders. 

Based on the causal graph in Figure~\ref{causal} (b) and the front-door adjustment, we introduce causal mechanism~\cite{glymour2016causal} to interpret these metric-based FSL methods, which provides a unified explanation. Specifically, we show that the existing metric-based FSL methods can reduce the spurious correlation that is defined in the causal graph in Figure~\ref{causal} (b). In deep learning, we can interpret the deep representation of the example as the mediator $Z$ between example $X$ and label $Y$, that is $X$ causes $Y$ only through $Z$. The true effect of $X$ on $Y$ can be identified via front-door adjustment. From this causal perspective, we find that existing metric-based FSL methods like Matching Networks and Prototypical Networks can well fit the intervened causal framework in some special cases of the front-door criterion~\cite{glymour2016causal} (please refer to Section \textbf{Causal Interpretation} for more details). 

Our causal interpretation can inspire us to design new causal methods for metric-based FSL. We find that previous metric-based methods mainly focus on the way how training examples interact with the intermediate variable $Z$ (the red line in Figure~\ref{causal} (b)), while little attention has been paid to the diversity of the representations of the example, i.e., $X \to Z$. Therefore, we further introduce causal few-shot methods to consider the diversity of representations $Z$. 
To better illustrate all this, we apply ensemble learning~\cite{dvornik2019diversity} and Bayesian  approximation~\cite{gal2015dropout} as two examples to obtain diverse representations. The ensemble method simply combines Tip-Adapter-F~\cite{radford2021learning} and Zero-shot BLIP~\cite{li2022blip} two models. And dropout~\cite{gal2015dropout} acts as a Bayesian approximation. These two simple causal methods are used as examples to show that our causal interpretation can provide a new perspective for designing FSL algorithms. 
In summary, our proposed methods are totally based on the formulation of the front-door adjustment. Compared to the existing metric-based methods, the proposed methods are more refined expression of the front-door adjustment. 
Experimental results have shown that our proposed methods can make significant improvements on various benchmark datasets. The proposed ensemble-based causal method improves 1-shot accuracy from 60.93\% to 68.19\% and 16-shot accuracy from 65.34\% to 72.73\% on ImageNet.

The contributions of our work are summarized as:

\begin{itemize}
    \item We formalize a way of FSL in causal framework. We find that existing mainstream metric-based  FSL methods can be well explained in the framework.
    \item We propose two causal methods as examples to deal with FSL problems, which consider both the interaction between examples and the diversity of representations. 
    \item We evaluate our methods on 10 benchmark few-shot image classification datasets and conduct ablation studies to explore their characteristics. Experimental results demonstrate the compelling performance of our methods across the board.
\end{itemize}

The remainder of this paper is organized as follows.  In Section 2, the related works are discussed. In Section 3, we will try to introduce a causal interpretation for several metric-based few-shot methods. This causal perspective could interpret the success of these metric-based methods. In Section 4, we further introduce two simple metric-based methods based on causal interpretation. In Section 5, we compare the proposed methods with several state-of-the-art baselines. Finally, Section 6 concludes this work.

\section{Related Work}

\subsection{Metric-Based Few-Shot Learning}

Metric-based FSL methods~\cite{liu2022learning,zhou2022forward,liu2022dynamic,li2021adaptive} aim to learn the metric space of examples. Siamase Neural Networks~\cite{koch2015siamese} regarded the FSL problem as a binary classification problem. Matching Networks~\cite{vinyals2016matching} proposed to compare training examples and test examples in a learned metric space. Prototypical Networks~\cite{snell2017prototypical} assigned test example to the nearest class prototype. Relation Network~\cite{sung2018learning} performed similarity computation via a neural network. BSNet~\cite{li2020bsnet} used a bi-similarity module to perform classification. CLIP-Adapter~\cite{gao2021clip} adopted a simple residual layer as a feature adapter over the feature extracted by pre-trained vision-language model CLIP~\cite{radford2021learning}. Tip-Adapter~\cite{zhang2022tip} improved CLIP-Adapter by effectively initializing the weight of the feature adapter.

Nevertheless, the success of these methods still lacks a solid theoretical explanation. In this paper, we interpret the success of metric-based methods from the perspective of causal mechanism. With the interpretation, we propose causal methods to deal with few-shot tasks.


\subsection{Causal Inference}

Causal inference~\cite{zhang2020causal, tang2020long, jiang2022role} denotes the pursuit of true causation instead of spurious correlation yielded by the confounder~\cite{glymour2016causal,huang2022deconfounded,mitrovic2021representation}. A general approach in machine learning is to maximize the observational conditional probability $P(Y|X)$ given example $X$ and label $Y$. Nevertheless, causal inference contends to deconfound~\cite{glymour2016causal} the training by using a new objective $P(Y|do(X))$ instead of $P(Y|X)$, which aims to find what $X$ truly causes $Y$. The $do$-operation denotes the pursuit of the causal effect of $X$ on $Y$ by intervening $X=x$. Two common deconfounding causal tools are used in causal inference: front-door adjustment~\cite{glymour2016causal} and back-door adjustment~\cite{glymour2016causal}. 

Causal inference was recently introduced to machine learning and applied to various tasks. For example, DIC~\cite{yang2021deconfounded} conducted image captioning by removing the spurious correlation yielded by the pre-training dataset via front-door adjustment and back-door adjustment. RED~\cite{huang2022deconfounded} was a confounder-agnostic visual grounding method, which aims to alleviate the effect of substitute confounder derived from observational data. ~\cite{mao2022causal} solved the out-of-distribution image classification problem in the way similar to front-door adjustment. ~\cite{yue2020interventional} proposed a FSL method from the view of causal mechanism, but their method was based on the assumption that the confounder is observable, which is not always the case as pointed out in ~\cite{li2021confounder}. Our methods are more general since we don't need to identify the confounder.

\subsection{Pre-trained Vision-Language Models}
Pre-trained vision-language models aim to improve the performance of downstream vision and language tasks by pre-training on large amounts of image-text pairs. Pre-trained vision-language models like ALBEF~\cite{li2021align}, BLIP~\cite{li2022blip}, and CLIP~\cite{gao2021clip} have shown amazing results on a wide range of downstream tasks even without fine-tuning. They learn to align the representations of images and texts in the way of contrastive representation learning~\cite{gao2021clip}, where an image and a text from the same pair are expected to have similar representations.


\section{Causal Interpretation}
In this paper, we consider the traditional $N$-way $K$-shot classification in FSL. For each example $x$, the successful FSL methods should learn its true effect on $y$, e.g., $bird's \ image \to bird$, but not the spurious correlations caused by the unobserved confounders, e.g., $sky \to bird$. 

To show why the existing FSL methods are successful, we first briefly review the front-door adjustment. 

\begin{figure}[!t]
    \centering
    \includegraphics[scale=0.2]{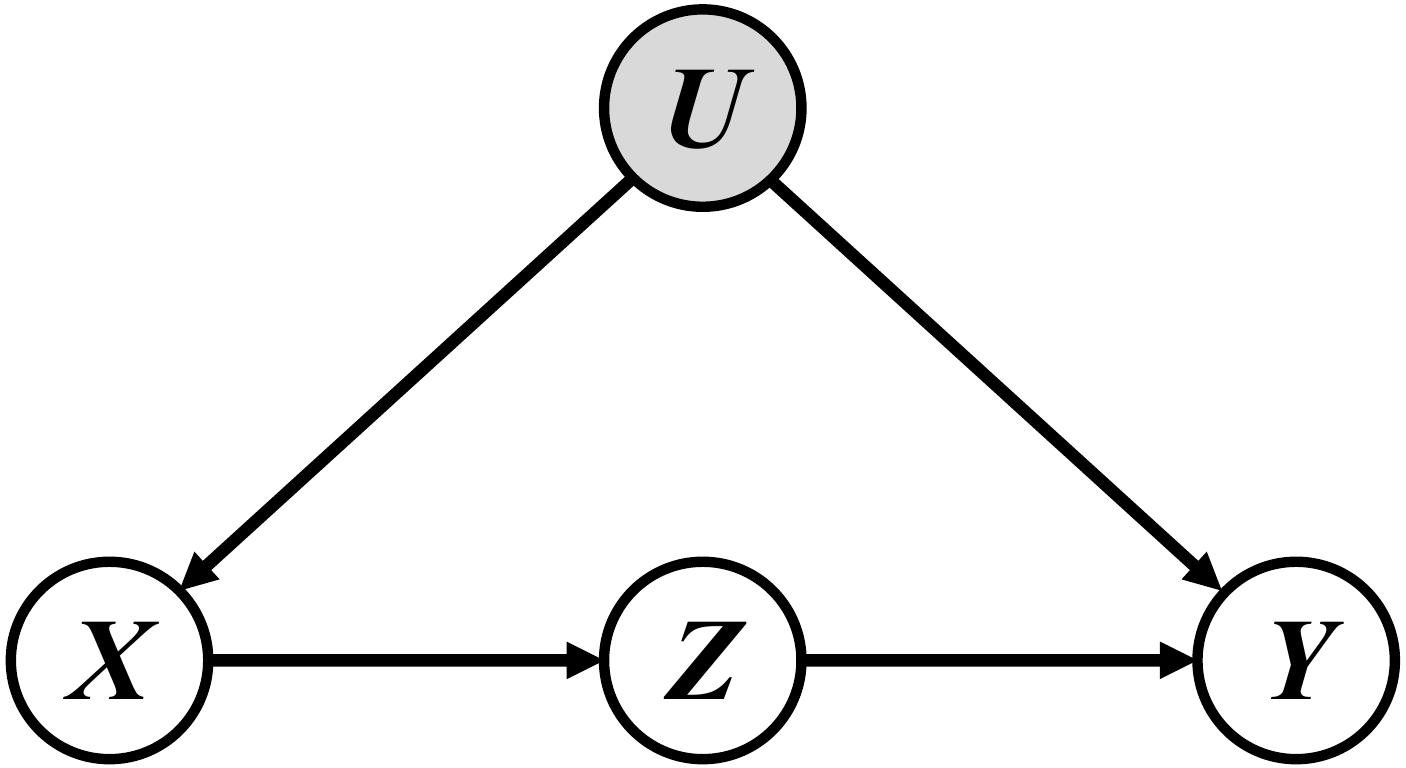}
    \caption{Illustration of front-door adjustment. $U$: unobserved confounder, $X$: example, $Z$: representation of the example, $Y$: label.}
    \label{frontDoor}
\end{figure}

\subsection{Background: Front-Door Adjustment}
Suppose that given a FSL task with a few training data $\{X, Y\}$, we aim to learn the true causal effect from example $X$ to label $Y$. Many metric-based deep methods~\cite{liu2022dynamic,zhou2022conditional} have been proposed for FSL. However, these deep methods always learn the co-existing associations between the training samples and labels, that is to maximize the observational conditional probability $P(Y|X)$. Since there are very few training examples and can not cover different contexts, e.g., all $K$ images show birds flying in the air, the model may learn the spurious correlations, e.g., the ``sky" and the label ``bird" are always co-existing and thus learn a spurious correlation $sky \to bird$ but not true causality. The existing work ~\cite{luo2021rectifying} also showed that the background information may mislead the classification in few-shot learning. The co-existing associations lack the ability to seek true causality, which could not generalize well to unseen examples.  
 
In this paper, we use a Structure Causal Model (SCM) ~\cite{radford2021learning} to formulate the causality of the few-shot recognition task. Consider the deep metric-based methods, given an input image $X$, it is firstly encoded into a feature representation $Z$ via the deep neural network. Then the feature $Z$ is used for classification $Y$. To capture the true causality, we also introduce the unobserved confounders $U$ into the causal graph. Hence, the causal graph is defined in Figure\ref{frontDoor}, where $X$ is an example, the mediator $Z$ is the feature representation of $X$, and $Y$ is the label.  $X \rightarrow Z \rightarrow Y$ represents that $X$ causes $Y$ only through the mediator $Z$. The $U$ is an unobserved confounder that influences both $X$ and $Y$, which can cause a spurious correlation between $X$ and $Y$. That is $X$ and $Y$ might be confounded by the unobserved confounders $U$. Specifically, the causal graph includes two paths: $X \to Y$ and $X \leftarrow U \to Y$.
 
 $\boldsymbol{X \to Y}$ indicates the true causality from example $X$ to label $Y$.
 
 $\boldsymbol{X \leftarrow U \to Y}$ indicates the path that causes spurious correlations between $X$ and $Y$. The path $U \to X$ (or $U \to Y$) denotes that unstable context $U$ determines the content of $X$ (or $Y$). For example, $U$ denotes ``taking pictures of birds in the sky''. Though the unstable context has no causal relationship with the label ``bird", the path $X \leftarrow U \to Y$ creates a spurious correlation between sky and bird.

In summary, simply maximizing the observational conditional probability $P(Y|X)$ can not always learn the true causality. The existence of confounders will mislead the training model to learn spurious correlations between the cause $X$ and the effect $Y$. Thus, the \textbf{do-operation} is introduced to find what $X$ truly causes $Y$: $\boldsymbol{P(y|do(x))}$.

With the causal graph, how to remove the spurious path and formulate the do-operation becomes a hot topic. \textbf{Front-door adjustment} is a deconfounding way in the causal mechanism, which is proposed to remove the spurious path $X \leftarrow U \to Y$ and find what $X$ truly causes $Y$. Here we just give a brief introduction of front-door adjustment,  please refer to~\cite{glymour2016causal} for more details of the front-door adjustment.

To remove the effect of the spurious path, the front-door adjustment introduces a new variable, the mediator $Z$, which is the representation of $X$. $X \rightarrow Z \rightarrow Y$ represents that $X$ causes $Y$ only through the mediator $Z$. For deep neural networks, $Z$ can be viewed as the deep feature of $X$. Thus the mediator $Z$ can be easily obtained. Then, the do operation is divided into two partial effects: 1) $P(z|do(x))$ and 2) $P(y|do(z))$.  
On the one side, for $x \in X$, $y \in Y$, and $z \in Z$, considering the partial effect of $X$ on $Z$, we have $P(z|do(x)) = P(z|x)$ since there is no backdoor path from $X$ to $Z$ (See Figure~\ref{frontDoor}). On the other side, considering the partial effect of $Z$ on $Y$, we have $P(y|do(z))={\mathbb E}_{P(x')}P(y|x', z)$ for $x' \in X$ (though $x$ and $x'$ both come from $X$, we distinguish them since $x$ is the input while $x'$ is used for back-door adjustment in the formula $P(y|do(z))$.  This is because the backdoor path from $Z$ to $Y$, namely $Z \leftarrow X \leftarrow U \rightarrow Y$, can be blocked by conditioning on $X$. Finally by chaining together the two partial effects $P(z|do(x))$ and $P(y|do(z))$, we have the following front-door adjustment formula~\cite{glymour2016causal}:
\begin{align}
    P(y|do(x)) 
    & = \sum_{z \in Z}[P(z|do(x))P(y|do(z))] \notag \\
    & = \sum_{z \in Z}[P(z|x){\mathbb E}_{P(x')}P(y|x', z)],
    \label{front-door1}
\end{align}
where $P(y|do(x))$ is the true effect of $x$ on $y$. With the do-operation $P(y|do(x))$, we can learn the causality and remove the spurious correlations caused by the path $X \leftarrow U \to Y$.

\subsection{Interpretation for Metric-Based FSL Methods}
With the above formulation, we show that we can re-formalize FSL methods and these methods can alleviate the effect of the unobserved confounders defined in Figure~\ref{causal} (b). The $U$ is the unobserved confounder that affects the distribution of example $X$ and label $Y$, which should be removed. Given an example $x$, a common procedure in deep learning is $x \rightarrow z \rightarrow y$~\cite{li2023adjustment}: using feature extractor $f$ to generate representation $z=f(x)$ at first. Then we use the representation $z$ to make prediction $y$, which indicates $x$ causes $y$ through the mediator $z$. Please note that the case in Figure~\ref{causal} (b) where representation serves as the mediator has been widely discussed and justified in recent works~\cite{li2021confounder,huang2022deconfounded,yang2021deconfounded}, where variable $Z$ satisfies the front-door criterion~\cite{glymour2016causal} between $X$ and $Y$. 

In the following, we will show that \textbf{the existing metric-based few-shot learning methods can be viewed as specific forms of front-door adjustment by setting 1) different values of $X'$, and 2) different interactive functions ${\mathbb E}_{P(x')} P(y|x',z)$ }. We then adopt the front-door adjustment to derive the intervened FSL framework. {Let $X'$ be a set of specific values, then we have the following approximation to \textbf{Eq.~(\ref{front-door1})}:
\begin{align}
    P(y|do(x)) \approx \sum_{z \in Z}[P(z|x)\sum_{x' \in X'}P(y|x',z)P(x')],
\label{approx}
\end{align}
where probability $P(y|x',z)$ can be interpreted as the probability $P(Y=y)$ predicted by exploring the relationship between the representation of example $x'$ and the representation $z$. The $P(x')$ is the probability of $x'$. 

Since the deep network $f$ is the deterministic mapping, the representation $f(x)$ of each example $x$ is unique. That is $x$ is deterministically mapped to $f(x)$. Thus we have $P(f(x)|x) = 1$. Using the Dirac delta function as the distribution of $Z$ in Eq.(2), the cardinality of representation set $Z$ is exactly one, i.e., $z=f(x)$, and we have $P(z|x) = 1$ as the same case of ~\cite{li2023adjustment}. Thus Eq. (2) can be further simplified as:
\begin{equation}
 P(y|do(x)) \approx \sum_{x' \in X'}P(y|x',z)P(x').
\end{equation}
where $z=f(x)$.

\subsubsection{Interpretation for Matching Networks}
In Matching Networks (MN)~\cite{vinyals2016matching}, the distance computation is performed for each pair of training example and test example. Let the labeled training set $S = \{(x_{i}',y_{i}')\}_{i=1}^{NK}$, given a test example $x$, MN computes a probability  over $y$ as follows:
\begin{equation}
    P_{\rm MN}(y|x,S) = \sum_{i=1}^{NK} a(x,x_{i}')y_{i}',
\label{matchOrigin}
\end{equation}
where $a$ is an attention mechanism that computes the similarity between two examples. Here $a$ can be viewed as a distance function and the predicted label is assigned according to the similarities among the training examples and test example.  

This can be interpreted as a special case by letting $X'$ be the training set and the interactive function $P(y|x',z)P(x') = a(z,x')y'$. Please note that $P(x')$ is to make sure the summing result is a valid conditional probability. We have:
\begin{align}
    P(y|do(x)) 
    &\approx P(z|x)\sum_{x' \in X'}P(y|x',z)P(x') \notag \\
    &= 1 \times \sum_{i=1}^{NK} a(z,x_{i}')y_{i}'  =  P_{\rm MN}(y|x,S),
\label{matchCausal}
\end{align}
where $a(z,x_{i}') = a(x,x_{i}')$ since $z$ is the unique representation of $x$ and $P(z|x)=1$.

According to \textbf{Eq.~(\ref{matchCausal})}, we can observe that MN is a method that can remove the effects of confounder by considering all examples in the training sets. It is possible to remove the effects of confounders defined in Figure~\ref{causal}. From our causal perspective, we can provide an explanation to interpret the success of Matching Networks.


\subsubsection{Interpretation for Prototypical Networks}
In Prototypical Networks(PN)~\cite{snell2017prototypical}, the distance computation is performed between test example and class prototypes $C$ of training examples. The $j$-th class prototype $c_j$ is defined as:
\begin{equation}
   c_j = \frac{1}{K} \sum_{(x',y'_j) \in s_j}f(x'), 
\end{equation}
where $s_j$ is the labeled training set of the $j$-th class, $f(x')$ is the representation of training example $x'$, and $y'_j$ is one-hot label. 

Given a test example $x$, the representation of the test example is also unique. PN computes classification probability as follows:
\begin{align}
   P_{\rm PN}(y|x,S) &= \sum_{j=1}^{N} a(x,c_j)y'_j. 
\label{protoOrigin}
\end{align}
where $a$ is an attention mechanism that computes the similarity and $y'_j$ is the $j$-th label. 

This can be interpreted as a special case by letting $X'$ be the set of class prototypes $C$ and the interactive function $P(y|x',z)P(x') = P(y|c_j,z)P(c_j) = a(z,c_j)y'_j$ in \textbf{Eq.~(\ref{approx})}:
\begin{align}
    P(y|do(x)) 
    & \approx P(z|x)\sum_{c_j\in C}P(y|c_j,z)P(c_j)  \notag \\
    &= 1 \times \sum_{j=1}^{N}a(z,c_j)y'_j = P_{\rm PN}(y|x,S),
\label{protoCausal}
\end{align}
where $z=f(x)$, $ C = \{c_1, \dots, c_N\}$, and $P(z|x)=1$. 

\subsubsection{Interpretation for CLIP/Tip-Adapter}
CLIP-Adapter~\cite{gao2021clip} further utilizes the relationship between the visual and textual representations, which learns the classification model with natural language supervision.  

Given a test example $x$, $f(x) \in \mathbb{R}^{1 \times D}$ is the visual representation extracted by the visual encoder, where $D$ is the dimension of $f(x)$. Let $W \in \mathbb{R}^{N \times D}$ be textual representations generated by embedding the descriptions/names of the classes. Each $w_j \in W$ ($j=1,\cdots,N$) represents one class. The logits of CLIP-Adapter(\textbf{Eq. (2)} in ~\cite{zhang2022tip}) can be rephrased as follows:
\begin{equation}
    {\rm logits_{CLIP}}(x) = \tau(f(x))W^T,
\end{equation}
where $\tau(f(x))$ is the updated representation of example $x$. 

This can also be interpreted as a special case in \textbf{Eq. (\ref{approx})}  by letting $X'$ be the set of textual representations $W$:
\begin{equation}
    P(y|do(x)) \approx P(z|x) \sum_{w_j \in W}P(y|w_j, z)P(w_j),
\end{equation}
where $z=\tau(f(x))$ and the interactive function is $\sum_{w_j \in W}P(y|w_j, z)P(w_j)=zW^T$.

In Tip-Adapter~\cite{zhang2022tip},  
classification is performed by summing up the predictions from two classifiers: 1) the weight of one classifier is derived from the labeled training set $S$, and 2) the weight of another classifier is derived from textual representations $W$. The logits of Tip-Adapter(\textbf{Eq. (6)} in ~\cite{zhang2022tip}) can be rephrased as:
\begin{equation}
    {\rm logits_{TIP}}(x) = \alpha  {\rm exp}(-Dis(x, X'))Y' + f(x)W^T,
\end{equation}
where $\alpha > 0$ is a hyper-parameter, $Dis$ is distance function, $X'/ Y'$ is the example/label matrix of labeled training set. 

This can be interpreted as a special case in \textbf{Eq. (\ref{approx})} by letting $X'$ be the training set $T$ and the set of textual representations $W$:
\begin{align}
    P(y|do(x)) \approx &P(z|x)\sum_{x' \in X'}P(y|x',z)P(x') \notag \\
    =&P(z|x) [\frac{\alpha}{\alpha + 1} \sum_{x' \in X'}P(y|x',z)P(x') +  \notag \\
    &
    \frac{1}{\alpha + 1}\sum_{x' \in X'}P(y|x',z)P(x')] \notag \\
    =&\frac{P(z|x)}{\alpha + 1} [\alpha \sum_{x' \in T}P(y|x',z)P(x') +  \notag \\
    &
    \sum_{w_j \in W}P(y|w_j,z)P(w_j) 
    ],
    \label{tip-adapter}
\end{align}

where $z=f(x)$ and the interactive functions are $\sum_{x' \in T}P(y|x',z)P(x')= {\rm exp}(-Dis(x, X'))Y'$ and $\sum_{w_j \in W}P(y|w_j,z)P(w_j) = zW^T$. The same as CLIP-Adapter, Tip-Adapter uses the ``class centroid'' $w$ in the textual domain provided by the textual encoder given the class name.

\begin{figure*}[t]
    \centering
     \includegraphics[scale=0.36]{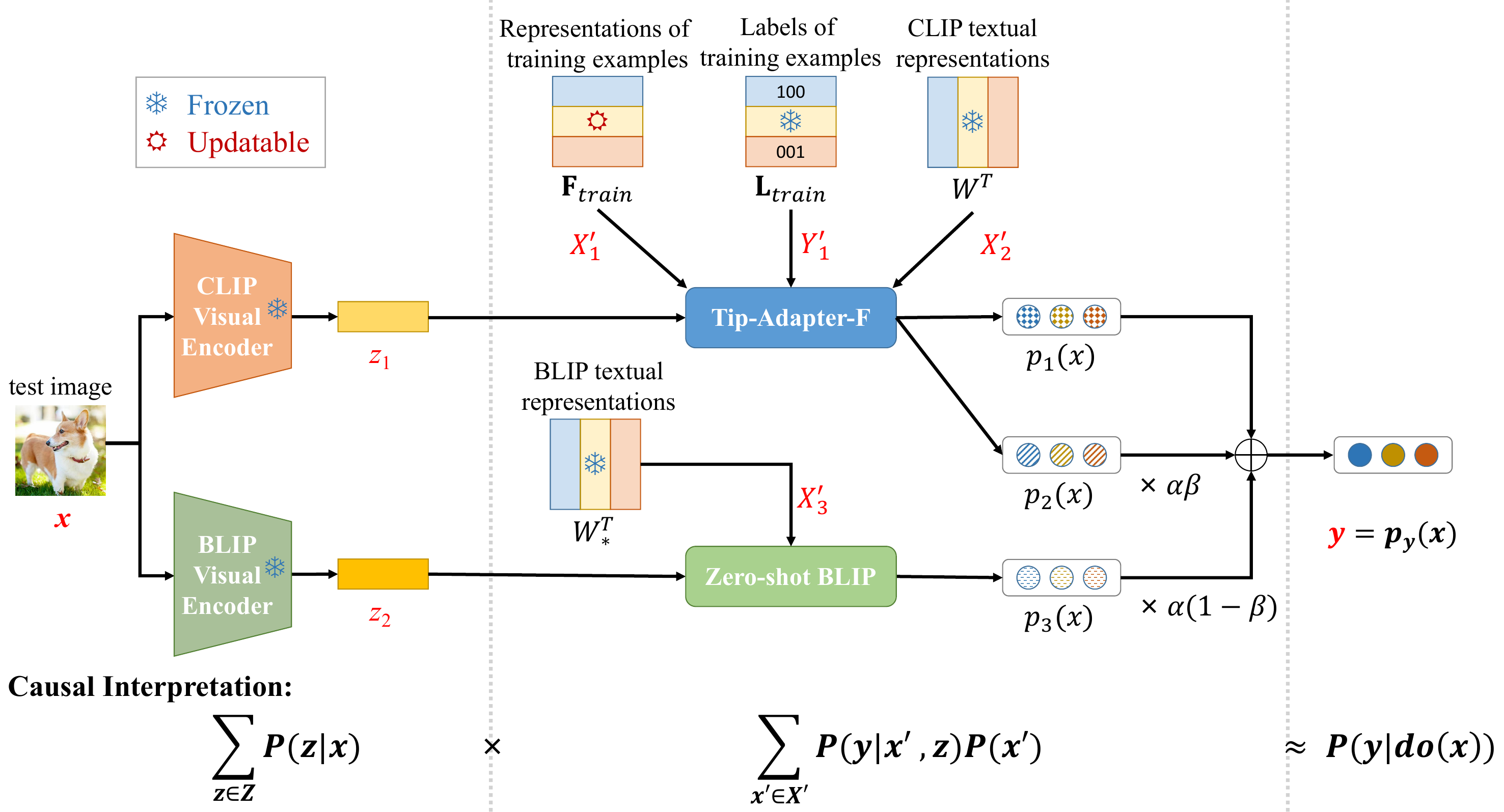}
    \caption{The pipeline of our ensemble method and its corresponding causal interpretation in \textbf{Eq. (\ref{approx})}. For a given test example $x$, we first obtain its representations $z_1$ and $z_2$ from diverse visual encoders. We then perform intermediate predictions via different models. The final prediction is the linear combination of intermediate predictions. Different parts of the pipeline and their corresponding causal interpretations are separated by dotted lines. Note that we only update representations ${\textbf{F}}_{train}$ as learnable parameters during training.}
    \label{pipeline}
\end{figure*}

\textbf{Remark 1: } \textit{Why do these methods work?} We introduce causal mechanism to interpret the success of these metric-based methods, where we reformulate their objectives according to the front-door adjustment formula. To further understand why these methods can alleviate the effects of confounders, we give intuitive examples.  

The first example is for Matching Networks (MN), which computes the distances between test example and the training set. Using these training examples may help to alleviate the effects of confounders. For instance, if we have other labeled training images (e.g., the plane in the sky) in the training set, it is possible to remove the spurious correlation $sky \to bird$ since ``sky" is not the distinguishable feature. 

The second example is for CLIP/Tip-Adapter. The textual information can be used to alleviate the effects of confounder, e.g., the word ``bird" involves no spurious correlation (e.g., $sky \to bird$) while the spurious correlation would exist in images under the unobserved confounder $U$. With the help of textual information, the spurious correlation $sky \to bird$ can also be reduced. 

\section{Two Causal Methods for FSL}
According to the above discussions, a general causal method for FSL that fully consider all aspects of the front-door formula can be derived from \textbf{Eq. (\ref{approx})} as
\begin{align}
    \underbrace{ \sum_{z \in Z}[P(z|x)}_{Representations} \underbrace{\sum_{x' \in X'}P(y|x',z)P(x')}_{Interactions}]. 
\end{align}
This causal perspective can guide us to design new algorithms for metric-based FSL. For example, the methods aforementioned mainly focus on the interactive functions, i.e., the term $P(y|x',z)P(x')$ in \textbf{Eq. (\ref{approx})}, but pay little attention to the diversity of representations, i.e., the term $\sum_{z \in Z} P(z|x)$. By summarization of diverse representations, we can remove the bias caused by different visual encoders, and therefore can better capture the causal correlations between the input and the target. In the following, we show that we can derive two causal methods based on our causal interpretation: the ensemble method and the stochastic mapping method. The ensemble method uses multiple visual backbones to obtain multiple representations; the stochastic mapping method generates representation in a stochastic way.


\subsection{Ensemble Method}

The pipeline of our ensemble method and its corresponding causal interpretation in \textbf{Eq. (\ref{approx})} are shown in Figure~\ref{pipeline}. To obtain diverse representations, we use the model Tip-Adapter-F~\cite{zhang2022tip}, which is the state-of-the-art model for FSL, along with the model Zero-shot BLIP~\cite{li2022blip}. Please note that other backbones and interactive functions are also applicable. After that, according to \textbf{Eq. (\ref{approx})}, the intermediate predictions are performed via Tip-Adapter-F and BLIP, and the final prediction is the linear combination of intermediate predictions. 

\subsubsection{Tip-Adapter-F}

 We follow Tip-Adapter-F~\cite{zhang2022tip} to consider the representations generated by CLIP backbone. For more details please refer to the original paper. It includes two interactive functions. One is to interact with the $N$-way $K$-shot labeled training samples as
\begin{align}
    p_1(x) = {\rm exp}( - (1 - z_1 { {\textbf{F}}^{T}_{train}} )){{\textbf{L}}_{train}},
\label{update}
\end{align}
where $z_1 = {\rm CLIPVisualEncoder}(x) $ is the feature of the test example $x$, ${\textbf{F}}_{train} ={\rm CLIPVisualEncoder}({\textbf{I}}_{train})$ are the features of all $NK$ training examples ${\textbf{I}}_{train}$ and ${\textbf{L}}_{train}$ denotes as their one-hot labels.

Another function is to interact with textual information as 
\begin{equation}
    p_2(x) = z_1 W^{T},
\label{wc}
\end{equation}
where the L2-normalized $D$-dimension textual representations $W$ for $N$ class names are extracted by CLIP textual encoder. The term $p_1(x)$ serves as new knowledge from few-shot tasks while the term $p_2(x)$ preserves the prior knowledge from CLIP model. 

\subsubsection{Zero-shot BLIP}
We then consider the representations generated by BLIP backbone. In this paper, we mainly focus on providing a new perspective instead of designing new technicalities for FSL. Thus we simply use the representation from BLIP as a complementary one.  

Given a test example $x$, we first get its L2-normalized $D$-dimension visual representation $z_2 $ from BLIP visual encoder:
\begin{equation}
    z_2 = {\rm BLIPVisualEncoder}(x).
\end{equation}

We then get the L2-normalized $D$-dimension textual representations $W_{*}$ for $N$ class names by BLIP textual encoder. The intermediate prediction $p_3(x)$ is calculated as follows:
\begin{equation}
    p_3(x) = z_2 W_{*}^{T}.
\label{wb}
\end{equation}

\subsubsection{Final Prediction}
The final prediction $p_y(x)$ is the linear combination of intermediate predictions mentioned above:
\begin{equation}
    p_y(x) = p_1(x) + \alpha [ \beta p_2(x) + (1-\beta) p_3(x)],
\label{final}
\end{equation}
where $\alpha$ and $\beta$ are hyper-parameters. Here $\alpha$ controls the trade-off between new knowledge from few-shot tasks and prior knowledge from pre-trained models while $\beta$ controls the trade-off between representations from CLIP backbone and BLIP backbone.

During fine-tuning, we only update representations ${\textbf{F}}_{train}$ as learnable parameters using cross-entropy loss while other parameters remain fixed. In other words, the gradients of ${\textbf{F}}_{train}$ will not be further back-propagated to the CLIP backbone. Specifically, the weights of $\textbf{L}_{train}$, CLIP backbone, and BLIP backbone remain fixed. The training objective of our method can be formulated as:
\begin{equation}
    \mathop{\arg\min}\limits_{{\textbf{F}}_{train}}\sum_{(x',y')\in S}CE(p_y(x'),y'),
\end{equation}
where $S$ is the labeled training set, $CE(.,.)$ is cross-entropy loss function. 

We don't fine-tune the whole model since fine-tuning large-scale vision-language model would lead to severe over-fitting~\cite{gao2021clip, zhang2022tip, sung2022vl}. Besides, the one-hot labels serving as ground-truth annotations should be kept frozen for anti-forgetting. 

\subsubsection{Causal Interpretation}
As indicated by \textbf{Eq. (\ref{approx})}, our method considers not only the relationship between examples but also the diversity of representations. We can rephrase our method as follows:
\begin{align}
        P(y|x) =& \sum_{z \in Z}[P(z|x)\sum_{x' \in X'}P(y|x', z)P(x')] \notag \\
           =& P(z_1 | x)\sum_{x' \in T}P(y|x', z_1)P(x') + \notag \\
            & P(z_1 | x)\sum_{w \in W}P(y|w, z_1)P(w) + \notag \\
            & P(z_2 | x)\sum_{w_* \in W_{*}}P(y|w_*, z_2)P(w_*),
\end{align}
where $z_1$ is the representation obtained from CLIP visual encoder, $z_2$ is the representation obtained from BLIP visual encoder, $T$ is training set, $W$ is provided by CLIP textual encoder, and $W_*$ is from the textual domain provided by BLIP textual encoder. 

Using diverse representations can remove the bias caused by visual encoders. Suppose that there are more images of ``birds in the sky'' in the training dataset of CLIP while there are more images of ``birds in the grass'' in the training dataset of BLIP. Then the using of multiple visual backbones can let the machine know that the background ``sky'' or ``grass'' is not the true cause of the label and therefore remove the effect of the unobserved confounder.

\textbf{Remark 2: } \textit{What is the contribution of the proposed method compared to  ensemble methods?}  Ensemble methods~\cite{jiang2021seen} combine multiple models to improve the accuracy. From this perspective, our method belongs to the ensemble methods. The main contribution is that we introduce causal mechanism to propose a novel way to guide the combining of multiple models. It also gives us a new perspective to understand these ensemble methods.  Compared to the existing ensemble method CaFo~\cite{Zhang_2023_CVPR} that encourages the similarity between representations (model with larger similarity gets larger ensemble weight as stated in \textbf{Eq. (9)} of its paper, the proposed method can better remove the effects of confounders (see Table~\ref{ensemble_results} for more details). 



\textbf{Remark 3: } The recent work~\cite{dvornik2019diversity} also conducts FSL by ensemble method, but its improvement is limited: it needs to fine-tune the whole model, which would lead to over-fitting on some datasets and the training is time-consuming for large-scale vision-language model~\cite{gao2021clip, zhang2022tip, sung2022vl}.

\subsection{Stochastic Mapping(SM) Method}
 The methods discussed above generate representations in a deterministic way. Such deterministic mapping is a one-to-one mapping, where one input is encoded as one representation. To obtain diverse representations, stochastic mapping as shown in Figure~\ref{sto}, which mines one-to-many relations among the input and representations, should be considered.  The typical work is using Bayesian networks~\cite{gal2015dropout}, which learns the distribution of weights instead of any specific weights. In this paper, we show that our causal interpretation can guide us to incorporate stochastic mapping~\cite{gal2015dropout} into the causal FSL method. 

\begin{figure}
    \centering
     \includegraphics[scale=0.33]{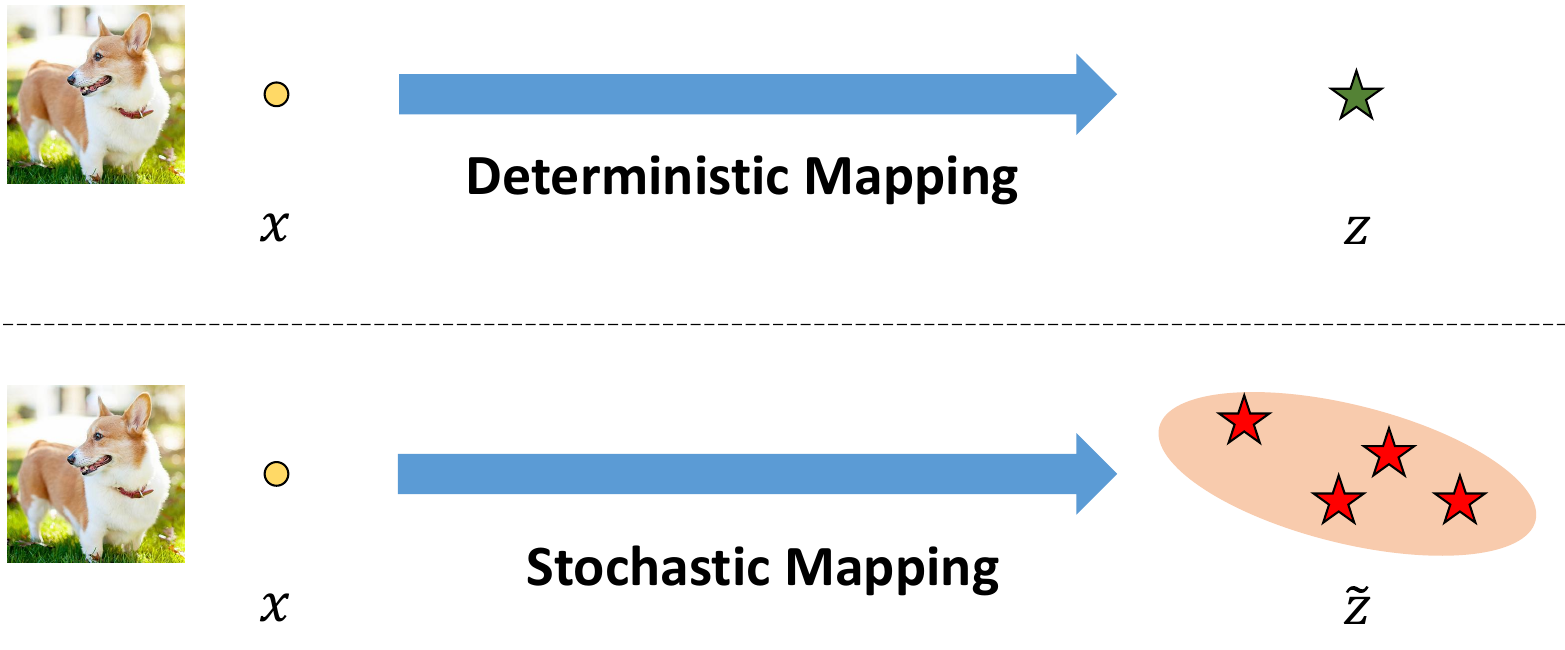}
    \caption{Comparison between the deterministic mapping and stochastic mapping. } 
    \label{sto}
\end{figure}

Since this paper does not focus on designing complex Bayesian algorithms, we take dropout as an example to show our main idea, where dropout has been proved mathematically equivalent to an approximation to a Bayesian model~\cite{gal2015dropout}. Specifically, we also use Tip-Adapter-F to demonstrate the stochastic mapping method. Let $z_1 = {\rm CLIPVisualEncoder}(x)$ denote the feature of the example $x$ extracted from the CLIP visual encoder. The dropped representation $\Tilde{z}$ with dropout probability $p$ is formulated as:
\begin{gather}
  b \sim {\rm Bernoulli}(1-p) \label{ber}  \\
    \Tilde{z} = b \circ z_1,
\end{gather}
where the dropout probability means that $p$ portion of $z_1$ is set to zero, and $\circ$ is the Hadamard product. After that, the following steps are the same as Tip-Adapter-F where the variable ${\textbf{F}}_{train}$ as learnable parameter is updated using cross-entropy loss while other parameters remain fixed. The dropout layer is removed during testing. 

\textbf{Causal Interpretation}: The causal interpretation of stochastic mapping is the same as the \textbf{Eq. (\ref{tip-adapter})}. The only difference is that the deterministic representation is replaced by multiple stochastic representations, where the $z$ is replaced with $\Tilde{z}$ and $P(z|x)$ is replaced with $\mathbb{E}_{\Tilde{z}} \ P(\Tilde{z}|x)$. 

Using dropout can also help the classification. The dropout has masked out the part of the image, and the machine is trained to  still predict correctly. This may let the machine learn what is true causality and therefore deconfound the training.

\textbf{Remark 4: } Please note that other Bayesian methods can also be incorporated into our causal interpretation. These two causal methods are just used as  examples to show our main idea.

\begin{table*}[ht]

\centering
\begin{tabular}{lccccc}
\hline
Models          &ImageNet           & Caltech101              & DTD                     & EuroSAT                     & FGVCAircraft   \\ \hline
Zero-shot BLIP  & 47.48             & 89.11                   & 47.18                   & 35.25                       & 5.52\\
Zero-shot CLIP  & 60.32             & 83.96                   & 41.33                   & 23.50                       & 16.20                            \\
Tip-Adapter     & 61.81             & 85.35                   & 47.02                   & 45.62                       & 20.13                            \\
Tip-Adapter-F   & 65.34             & 90.89                   & 67.29                   & 78.25                       & 32.76                  \\ \hline
Ours-Ensemble & \textbf{72.73} & \textbf{96.14}  & \textbf{69.20} & \textbf{82.25}  & 30.54              \\
Ours-SM & 64.13  &  \textbf{92.08}   &  \textbf{68.09}  & \textbf{81.50}  & \textbf{37.53}
\\ \hline
\\
\hline
Models          & Flowers102     & Food101        & OxfordPets   & StanfordCars   & SUN397         \\ \hline
Zero-shot BLIP  & 57.11          & 71.81          & 59.80          & 71.06          & 45.57          \\
Zero-shot CLIP  & 62.99          & 76.93          & 83.05          & 54.37          & 58.95          \\
Tip-Adapter     & 68.87          & 78.08          & 84.08          & 60.45          & 64.11          \\
Tip-Adapter-F   & 91.42          & 81.31          & 88.58          & 72.86          & 71.14          \\   \hline
Ours-Ensemble & \textbf{94.36} & \textbf{86.17} & \textbf{90.54} & \textbf{84.95} & \textbf{74.85}              \\
Ours-SM &  \textbf{95.59}    &  80.42     & 88.39    & \textbf{75.95}   & \textbf{71.48}

\\ \hline
\end{tabular}
\caption{Accuracy (\%) of different models on various datasets. Methods are evaluated with 16 shots~\cite{zhou2022conditional}.}
\label{datasets}
\end{table*}

\section{Experimental Results}
In this section, we evaluate our methods w.r.t. classification accuracy on various datasets compared with several state-of-the-art baselines including Zero-shot BLIP~\cite{li2022blip}, Zero-shot CLIP~\cite{gao2021clip}, Tip-Adapter~\cite{zhang2022tip}, and Tip-Adapter-F~\cite{zhang2022tip}. Besides, we also conduct ablation studies to explore the characteristics of our methods. 

\subsection{Datasets}
We conduct experiments on 10 benchmark image classification datasets: 
\begin{itemize}
    \item ImageNet~\cite{deng2009imagenet} is an image dataset organized according to the WordNet hierarchy, which includes 1,000 object classes.

    \item Caltech101~\cite{fei2004learning} consists of  101 object categories, and each object category contains about 40 to 800 images. 

    \item DTD~\cite{cimpoi2014describing} aims to seek the texture representation for recognizing describable texture attributes in images, which consists of 5,640 texture images with 47 attributes. 

    \item EuroSAT~\cite{8736785} is a dataset for land use and land cover classification, which 13 spectral bands and 27,000 labeled images. 

    \item FGVCAircraft~\cite{maji2013fine} is a fine-grained dataset of aircraft, which contains 10,000 images and 100 aircraft models. 

    \item Flowers102~\cite{nilsback2008automated} consists of 102 flower categories, each flower category contains between 40 and 258 images.

    \item Food101~\cite{bossard2014food}  consists of 101 food categories. Each category contains 750 training and 250 test images.

    \item OxfordPets~\cite{parkhi12a} is a fine-grained dataset, which contains 37 different breeds of cats and dogs.

    \item StanfordCars~\cite{Krause_2013_ICCV_Workshops} consists of 196 classes of cars with a total of 16,185 images.

    \item SUN397~\cite{xiao2010sun} is a dataset for scene categorization, which includes 899 categories and 130,519 images. The 397 well-sampled categories are used to evaluate.   
\end{itemize}
For each dataset, we follow the settings of~\cite{gao2021clip,zhang2022tip} and sample images to construct two disjoint sets: training set and test set. The number of training examples is very small, and the training set includes 1, 2, 4, 8, or 16 examples. Each method is trained on training set and tested on test set. 

\subsection{Implementation Detail}
For the pre-trained CLIP backbone, we use ResNet-50~\cite{he2016deep} as the visual encoder and transformer~\cite{vaswani2017attention} as the textual encoder. For the pre-trained BLIP backbone, we use ViT-Large~\cite{dosovitskiy2020image} with $16 \times 16$ patch size as the visual encoder and BERT~\cite{devlin2018bert} as the textual encoder. The prompt design of CLIP backbone and the way of image pre-processing follows ~\cite{zhang2022tip}.

 We use AdamW~\cite{loshchilov2018decoupled} as optimizer and the learning rate is set to $10^{-3}$. The batch size is set to 256 for the training set. The number of shots, or the size of the training set, is set to 16 as default, which follows the settings of the existing few-shot methods~\cite{zhou2022conditional}. For Ensemble Method, the hyper-parameter $\alpha$ is set to 100 and $\beta$ to 0.5 in \textbf{Eq. (\ref{final})} by default, the number of training epochs is 20. For SM Method, the dropout probability $p$ is set to 0.5 in \textbf{Eq. (\ref{ber})} by default, the number of training epochs is 80.

\subsection{Comparison on Various Datasets}

Table~\ref{datasets} shows the performance of our methods on 10 benchmark image classification datasets with 16 shots. The CLIP visual backbone is ResNet-50 and the BLIP visual backbone is ViT-Large with $16 \times 16$ patch size for all methods. As Table~\ref{datasets} shows, our methods can greatly improve classification accuracies on various datasets, which are increased by at least 1.91\% on all benchmark image classification datasets except FGVCAircraft for Ensemble Method. The p-value
results, which are based on the paired Wilcoxon signed rank
test~\cite{woolson2007wilcoxon} across the 10 datasets, indicate the significant performance improvement of Ensemble Method over all four baselines
at the 99.5\% confidence level; SM Method over Tip-Adapter-F at the 95\% confidence level and over other three baselines at the 99.5\% confidence level. Accuracy of Ensemble Method is marginally lower than that of Tip-Adapter-F on FGVCAircraft. We attribute the degraded performance on FGVCAircraft to the lack of generalization of BLIP on FGVCAircraft since Zero-shot BLIP only achieves 5.52\% accuracy on FGVCAircraft. The accuracy of SM Method can also boost the performance for most cases, for example, the accuracy is increased by 4.77\% on FGVCAircraft.

We also compare methods with various shots for various datasets and the results are reported in Figure \ref{shotsv}, which shows that our Ensemble Method and SM Method can boost the performance of Tip-Adapter-F on most datasets like Flowers102 and StanfordCars.

\begin{figure*}[h]
    \centering
  \includegraphics[scale=1.5]{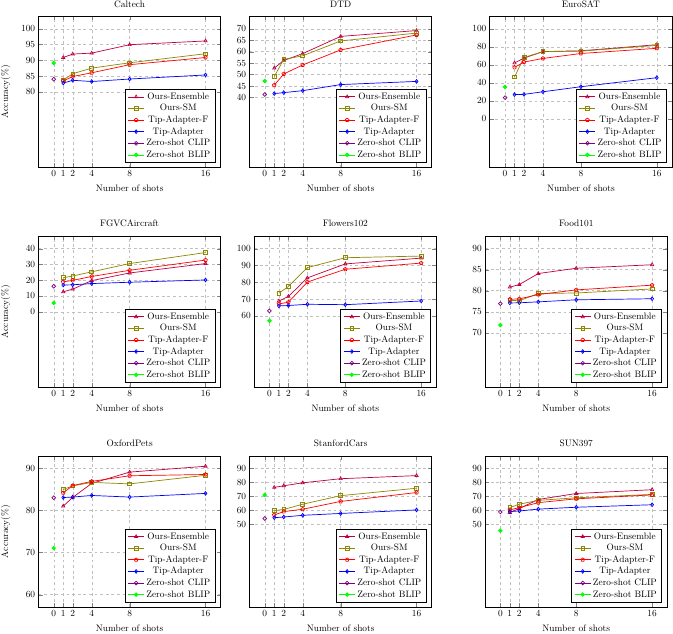}
    \caption{Accuracy(\%) of different models with various shots on various datasets. (Best viewed in color.)}    
     \label{shotsv}
\end{figure*}


\begin{figure}[ht]
    \centering
   \includegraphics[scale=0.9]{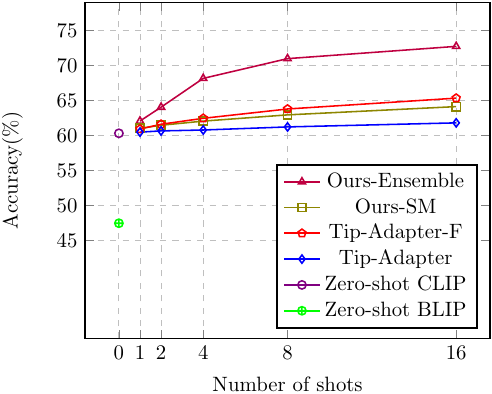}
    \caption{Accuracy(\%) of different models with various shots on ImageNet. (Best viewed in color.)}
    
    \label{shots}
\end{figure}
As Figure \ref{shots} shows, our Ensemble Method can achieve significant improvement on ImageNet with various shots. It surpasses all previous methods in all few-shot settings on ImageNet, which demonstrates the superiority of using diverse representations. 

\subsection{Results with Various CLIP Visual Encoders}

\begin{table}[h]

\centering
\begin{tabular}{lcccc}
\hline
Models         & RN50           & RN101          & ViT/32         & ViT/16         \\ \hline
Zero-shot CLIP & 60.32          & 62.53          & 63.81          & 68.74          \\
Tip-Adapter    & 61.81          & 64.01          & 65.34          & 70.30          \\
Tip-Adapter-F  & 65.34          & 68.48          & 68.52          & 73.68          \\ 
Ours-Ensemble         & \textbf{72.73} & \textbf{73.50} & \textbf{73.24} & \textbf{75.16} \\ 
Ours-SM &64.13 &67.60 &67.76 &73.21 \\
\hline
\end{tabular}
\caption{Accuracy (\%) of different models with various CLIP visual encoders. RN50 refers to ResNet-50 while ViT/32 refers to ViT-Base with 32 × 32 patch size. }
\label{clipChange}
\end{table}

We test our model with various CLIP visual encoders on ImageNet and the number of shots is set to 16. We adopt ResNet~\cite{he2016deep} and ViT~\cite{dosovitskiy2020image} for comparison. As Table \ref{clipChange} shows, our Ensemble Method can gain significant improvement with various CLIP visual encoders. 


\subsection{Comparison to Ensemble Methods}

Ensemble methods~\cite{jiang2021seen} involve combining predictions from various classifiers. First, for a fair comparison, we implement the baseline ensemble method by averaging the predictions from  CLIP and  BLIP two backbones, which are the same backbones as our method.  Second, we also compare with CaFo~\cite{Zhang_2023_CVPR}. CaFo is a recently proposed ensemble method for few-shot learning, which incorporates knowledge of various pre-training paradigms, including CLIP~\cite{gao2021clip}, DINO~\cite{caron2021emerging}, GPT-3~\cite{brown2020language}, and DALL-E~\cite{ramesh2021zero}. 

The comparison between our method and the ensemble methods is shown in Table~\ref{ensemble_results}. According to Table~\ref{ensemble_results}, we can see that our ensemble method derived from causal perspective also performs better than other SOTA ensemble methods with more backbones. Please note that CaFo uses GPT-3, CLIP, DINO and DALL-E four very large models. Even so, our simple causal method also has a competitive performance compared to CaFo. In summary, our proposed method can be viewed as an extension of the baseline ensemble method by fully considering all aspects of the front-door formula, especially the relationship between examples, which can better deconfound the confounder and boost the performance a lot. 

\begin{table*}[th]
\centering
\begin{tabular}{lccccc}
\hline
Models          &ImageNet           & Caltech101              & DTD                     & EuroSAT                     & FGVCAircraft   \\ \hline
CaFo~\cite{Zhang_2023_CVPR} (CLIP+DINO+GPT-3+DALL-E)  & 68.79             & 94.60                   & \textbf{69.62}                   & \textbf{88.68}                       & \textbf{49.05}                  \\
Baseline Ensemble (CLIP+BLIP) & 60.90    & 91.09      & 48.56 & 39.75   & 10.71        \\
\hline
Ours-Ensemble (CLIP+BLIP) & \textbf{72.73} & \textbf{96.14}  & 69.20 & 82.25  & 30.54            
\\ \hline
\\
\hline
Models          & Flowers102     & Food101        & OxfordPets   & StanfordCars   & SUN397         \\ \hline
CaFo~\cite{Zhang_2023_CVPR} (CLIP+DINO+GPT-3+DALL-E)  & \textbf{95.86}          & 79.30          & \textbf{91.55}          & 76.73          & 72.60          \\   
Baseline Ensemble (CLIP+BLIP) & 62.75          & 79.15          & 77.41          & 74.90          & 57.12          \\
\hline
Ours-Ensemble (CLIP+BLIP) & 94.36 & \textbf{86.17} & 90.54 & \textbf{84.95} & \textbf{74.85}              
\\ \hline
\end{tabular}
\caption{Comparison to ensemble models on various datasets. Methods are evaluated with 16 shots.}
\label{ensemble_results}
\end{table*}

\subsection{Comparison to other Prompt-based Methods}
In this set of experiments, we compare two state-of-the-art prompt-based methods, i.e., ProGrad~\cite{zhu2022prompt} and WiSE-FT(LC)~\cite{Wortsman_2022_CVPR}. The prompt-based methods  utilize the pre-trained  vision-language models and can show a fast adaptation to downstream tasks.

Table~\ref{promptbase_results} shows the comparison results. We can see that our proposed two simple methods can also perform better than the existing prompt-based methods although they can fully utilize large-scale pre-trained models like CLIP. For example, the accuracy of ProGrad is 63.45\% on ImageNet dataset, and our methods are 72.73\% and 64.13\%. On DTD dataset, our SM method achieves 68.09\% compared to 63.87\% of ProGrad and 54.20\% of WiSE-FT. The results indicate that it is beneficial to consider the causal interpretation into the designs of few-shot methods.  

\begin{table*}[th]
\centering
\begin{tabular}{lccccc}
\hline
Models          &ImageNet           & Caltech101              & DTD                     & EuroSAT                     & FGVCAircraft   \\ \hline
WiSE-FT(LC)~\cite{Wortsman_2022_CVPR} & 51.17             & 80.89                   & 54.20                   & 66.12                       & 18.72                            \\
ProGrad~\cite{zhu2022prompt}  & 63.45             & 92.10                   & 63.87                   & \textbf{83.29}                       & 30.25\\
\hline
Ours-Ensemble & \textbf{72.73} & \textbf{96.14}  & \textbf{69.20} & 82.25  & 30.54   \\
Ours-SM & 64.13  &  92.08   &  68.09  & 81.50  & \textbf{37.53}
\\ \hline
\\
\hline
Models          & Flowers102     & Food101        & OxfordPets   & StanfordCars   & SUN397         \\ \hline
WiSE-FT(LC)~\cite{Wortsman_2022_CVPR} & 77.94          & 69.45          & 66.18          & 51.72          & 60.67          \\
ProGrad~\cite{zhu2022prompt} & 94.37          & 78.41          & 89.00          & 73.46          & 69.84          \\
\hline
Ours-Ensemble & 94.36 & \textbf{86.17} & \textbf{90.54} & \textbf{84.95} & \textbf{74.85}              
\\
Ours-SM &  \textbf{95.59}    &  80.42     & 88.39    & \textbf{75.95}   & \textbf{71.48} \\
\hline
\end{tabular}
\caption{Comparison to prompt-based models on various datasets. Methods are evaluated with 16 shots.}
\label{promptbase_results}
\end{table*}

\subsection{Ablation Studies}
In this set of experiments, we conduct ablation studies on ImageNet to analyze the characteristics of our methods.  All experiments are tested with 16 shots.

\subsubsection{Analysis of Different $\alpha$ and $\beta$}

\begin{table}[]
\centering
\begin{tabular}{|c|ccccc|}
\hline
\diagbox{$\alpha$}{$\beta$}      & 0.1   & 0.3   & 0.5            & 0.7   & 0.9   \\ \hline
1     & 55.90 & 55.83 & 55.70          & 55.61 & 55.47 \\
10    & 62.30 & 61.39 & 60.16          & 59.09 & 57.86 \\
100   & 69.51 & 71.38 & \textbf{72.73} & 72.42 & 68.92 \\
1000  & 64.17 & 67.14 & 69.61          & 70.43 & 67.29 \\
10000 & 60.77 & 64.14 & 67.26          & 68.88 & 66.21 \\ \hline
\end{tabular}
\caption{Grid search of our Ensemble Method varying $\alpha$ and $\beta$.}
\label{grid}
\end{table}

As indicated by \textbf{Eq. (\ref{final})}, hyper-parameter $\alpha$ controls the trade-off between new knowledge from few-shot tasks and prior knowledge from pre-trained model. Therefore, as $\alpha$ becomes smaller, our model tends to learn more from the current few-shot task and less otherwise. 

As \textbf{Eq. (\ref{final})} shows, we can control the influence of CLIP backbone and BLIP backbone by changing the value of hyper-parameter $\beta$. If $\beta$ is a larger value, more information is depended by our model from the CLIP backbone.

We perform grid search and $\alpha$ is varied from 1 to 10000 and $\beta$ is from 0.1 to 0.9. As Table \ref{grid} shows, our model performs best when $\alpha$ is a moderate value 100. This implies that we should strike a good balance between new knowledge from few-shot tasks and prior knowledge from the pre-trained model. our model performs best when $\beta$ is a moderate value 0.5. This indicates that the CLIP backbone and the BLIP backbone contribute approximately equally to the final prediction. We choose $\alpha = 100$ and $\beta = 0.5$ as our setting after grid search.

\subsubsection{Analysis of Different Intermediate Predictions}
We further conduct ablation study to explore the effect of each intermediate prediction on the final prediction w.r.t. \textbf{Eq. (\ref{final})} for Ensemble Method. We use a tick ``$\checked$'' to indicate that we adopt the intermediate prediction for the final prediction. Here we set $\alpha$ to 100 and $\beta$ to 0.5 if they appear as the coefficient for the final prediction. Note that during fine-tuning we only update the representations ${\textbf{F}}_{train}$ in \textbf{Eq. (\ref{update})} if possible. Prediction $p_1$ alone is essentially the same as Matching Networks while few parameters can be updated, prediction $p_2$ alone represents Zero-shot CLIP, and prediction $p_3$ alone represents Zero-shot BLIP. $p_1$ with $p_2$ represents Tip-Adapter-F under our hyper-parameter setting. $p_1$ along with $p_2$ and $p_3$ represents our proposed method. As Table \ref{intermediate} shows, prediction from diverse representations is always better than that from a single representation, which strongly demonstrates the necessity for considering the diversity of representations. 

\subsubsection{Analysis of Different $p$}
We've also conducted experiments to explore the influence of the dropout probability $p$ in \textbf{Eq. (\ref{ber})}. When $p$ is smaller, the representation becomes more deterministic during training; otherwise the representation becomes more stochastic. The value $p$ is varied from 0.1 to 0.9. As Table \ref{tunep} shows, our SM Method performs well when $p$ is close to 0.5.


\begin{table}[h]

\centering
\begin{tabular}{cccc}
\hline
$p_1$   & $p_2$   & \multicolumn{1}{c|}{$p_3$}   & Accuracy(\%)  \\ \hline
\checked    &      & \multicolumn{1}{c|}{}     & 55.21     \\
     & \checked    & \multicolumn{1}{c|}{}     & 60.32     \\
     &      & \multicolumn{1}{c|}{\checked}    & 47.48     \\
\checked    & \checked    & \multicolumn{1}{c|}{}     &  64.72    \\
\checked    &      & \multicolumn{1}{c|}{\checked}    &  70.50   \\
     & \checked    & \multicolumn{1}{c|}{\checked}    & 63.19     \\
\checked    & \checked    & \multicolumn{1}{c|}{\checked}    & 72.73     \\ \hline
\end{tabular}
\caption{Ablation studies of our Ensemble Method with different intermediate predictions. We use a tick ``$\checked$'' to indicate that the intermediate prediction is adopted for the final prediction.}
\label{intermediate}
\end{table}


\begin{table}[h]

\centering
\begin{tabular}{cccccc}
\hline
$p$               & 0.1   & 0.3  & 0.5   & 0.7  & 0.9            \\
Accuracy(\%)         & 62.90 &  63.53 & 64.13 & 64.60 & 62.84        \\ \hline
\end{tabular}
\caption{Ablation studies of our SM Method varying $p$.}
\label{tunep}
\end{table}

\section{Conclusion}
In this paper, we interpreted existing mainstream metric-based few-shot learning methods from the perspective of causal mechanisms. We found that previous metric-based methods mainly focused on the way how examples interact but paid little attention to the diversity of the representations. Therefore, inspired by the front-door adjustment formula, we proposed causal methods to tackle few-shot tasks considering not only the relationship between examples but also the diversity of representations. Conducted experimental results have demonstrated the superiority of our proposed methods. In our future work, we aim to give a better theoretical explanation for more few-shot learning methods.


%

\ifCLASSOPTIONcaptionsoff
  \newpage
\fi



%

{\small
\bibliographystyle{ieee_fullname}
\bibliography{egbib}
}
%

\begin{IEEEbiography}[{\includegraphics[width=1in,height=1.25in,clip,keepaspectratio]
{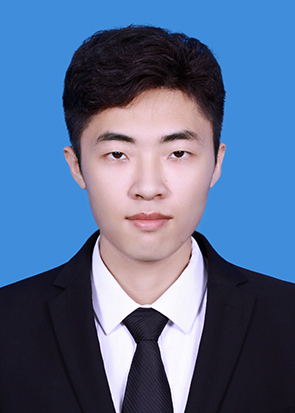}}]{Guoliang Lin}
received his B.S. degree from Sun Yat-sen University in 2021. He is currently working toward the M.S. degree with School of Computer Science and Engineering, Sun Yat-Sen University. His research interests include incremental learning, few-shot learning and causal inference.
\end{IEEEbiography}

\begin{IEEEbiography}[{\includegraphics[width=1in,height=1.25in,clip,keepaspectratio]{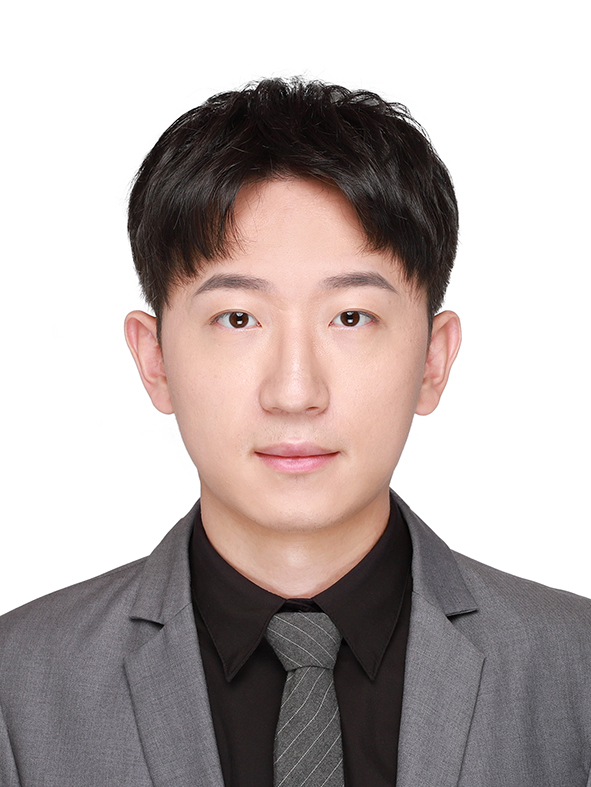}}]
{Yongheng Xu}
received his B.S. degree from Sun Yat-sen University in 2023. He is currently working toward the Ph.D. degree with School of Computer Science and Engineering, Sun Yat-Sen University. His research interests include incremental learning and few-shot learning.
\end{IEEEbiography}

\begin{IEEEbiography}[{\includegraphics[width=1in,height=1.25in,clip,keepaspectratio]{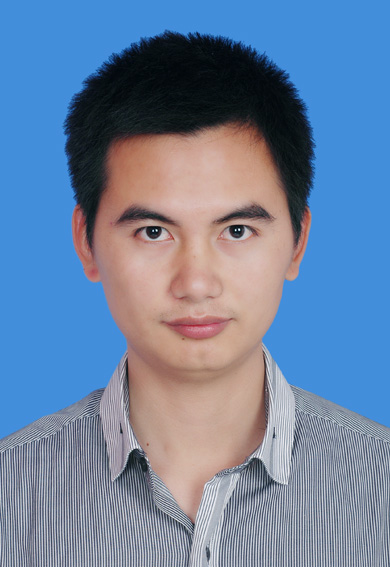}}]{Hanjiang Lai}
	received his B.S. and Ph.D. degrees from Sun Yat-sen University in 2009 and 2014, respectively. He worked as a research fellow at the National University of Singapore during 2014-2015. He is currently an associate professor at Sun Yat-Sen University. His research interests include machine learning algorithms, deep learning, and data mining.
\end{IEEEbiography}

\begin{IEEEbiography}[{\includegraphics[width=1in,height=1.25in,clip,keepaspectratio]{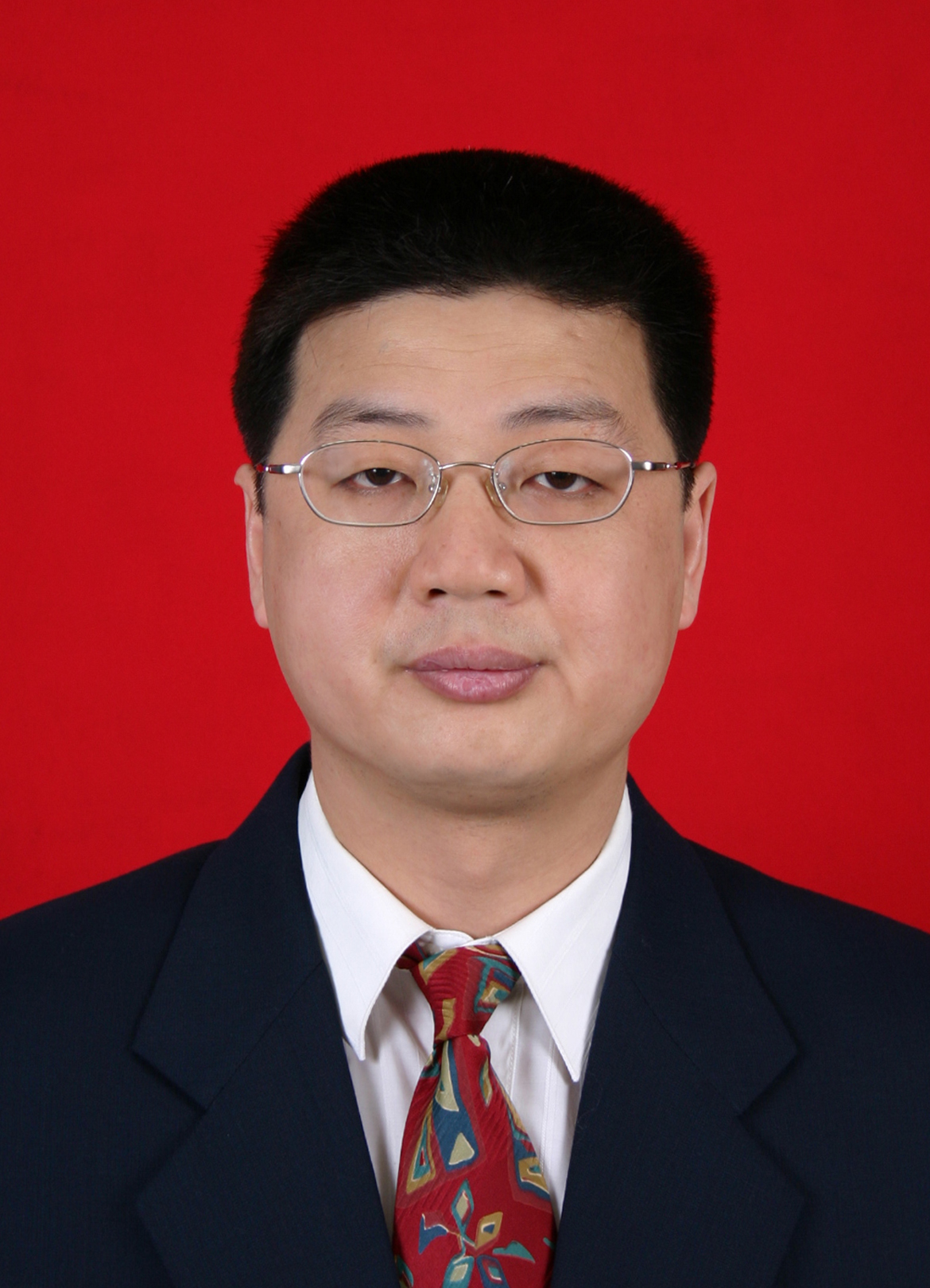}}]{Jian Yin}
 received the B.S., M.S., and Ph.D. degrees from Wuhan University, China, in 1989, 1991, and 1994, respectively, all in computer science. He joined Sun Yat-Sen University in July 1994 and now he is a professor at School of Artificial Intelligence. He has published more than 200 refereed journal and conference papers. His current research interests are in the areas of Data Mining, Artificial Intelligence, and Machine Learning. He is a senior member of China Computer Federation.
\end{IEEEbiography}





\end{document}